# AN ADAPTIVE GMM APPROACH TO BACKGROUND SUBTRACTION FOR APPLICATION IN REAL TIME SURVEILLANCE

**Subra Mukherjee[1], Karen Das[2]**

*Department of Electronics and telecommunication Assam Don Bosco University, Guwahati, India*
*subra_mukherjee@yahoo.in, karenkdas@gmail.com*

**Abstract**
*Efficient security management has become an important parameter in today's world. As the problem is growing, there is an urgent need for the introduction of advanced technology and equipment to improve the state-of art of surveillance. In this paper we propose a model for real time background subtraction using AGMM. The proposed model is robust and adaptable to dynamic background, fast illumination changes, repetitive motion. Also we have incorporated a method for detecting shadows using the Horpresert color model. The proposed model can be employed for monitoring areas where movement or entry is highly restricted. So on detection of any unexpected events in the scene an alarm can be triggered and hence we can achieve real time surveillance even in the absence of constant human monitoring.*

***Keywords****-Background subtraction, Adaptive Gaussian Mixture model (AGMM), surveillance, Horpresert color model.*

---

## I. INTRODUCTION

The present day video surveillance system has two main drawbacks: firstly they are not adaptable to different operative scenarios (they only work for a well known structured model). Secondly they need a human assistance to identify and label a specific event [2].Moreover dependency on human beings for a threat alert cannot be fully trusted. The general surveillance cameras are like machines that can only see, but cannot decide or identify things or events by itself. So, keeping in mind the present day scenario, it is important that we make our surveillance system intelligent and smart. For any real time surveillance the first step would be to efficiently detect motion and then extract the region of interest (ROI). For this, three main processes used are: optical flow, frame differencing and background subtraction [3].Optical flow can be used for detection even without any prior knowledge of the background and works well even when the camera is moving. However it is very sensitive to noise and requires special hardware. Frame differencing can detect target in dynamically changing background, but the segmentation of target is not integrated. And background subtraction is one of the most widely used methods for segmentation of dynamic scene from a video. A large number of methods and techniques have been cited in literature for background subtraction, each of them differing by the model or type of parameters used. However most of the models have a difficulty when dealing with bimodal background, sudden changes in illumination, frequent repetitive motions. In this paper, we model the background using the GMM approach proposed by Stauffer and Grimson [1]. The proposed model can be used to detect any moving object in the scene. Moreover it does not need any predefined background model in the beginning. It adaptively reads the video and any scene that is constant or static beyond a certain time is again considered as background. Moreover it responds well to dynamically changing background and even when the camera is moving. Shadows often encroach upon the foreground during segmentation and often causes problem in correct detection of object. For application in surveillance it is very important to correctly detect shadows from the extracted foreground. Therefore we have presented a shadow detection technique using the horpresert color model for moving object detection.

## II. RELATED WORK

Background modeling is generally done by analyzing some of the regular statistical characteristics and then object is detected by comparing the current frame with the modeled background. However, though it sounds simple, but this technique seems to be inadequate when dealing with complex environment. Most of the available non-adaptive methods of background subtraction cannot be used in surveillance because it requires manual initialization. And in surveillance, it is not possible to have a pre-defined set of background model before-hand. So, it is very important that the background model is adaptive and robust. Many background subtraction methods have been proposed in the past decades including Running Gaussian Average, Temporal Median Filter, Mixture of Gaussians, Kalman Filter and Co-occurrence of Image Variations. However the mixture model is widely used by researchers for application in surveillance.

A variety of methods have been proposed for modeling the background. Initially a single Gaussian was used to find the variances of the pixel intensity levels in the image sequences [4]. However generally multiple surfaces appear in a pixel and so each pixel should be represented by a mixture of Gaussians.

The GMM method was firstly used to model the pixel process[5] by N.Friedman et al. Stauffer & Grimson [1] modeled the values of a particular pixel as a mixture of Gaussians and based on the persistence and variance of each pixel of the Gaussians of the mixture, they determined which pixels corresponds to background colors. They also used an online approximation to update the model. Another similar work was done in [6]. They proposed an algorithm for modeling background which could increase or decrease the number of Gaussian components based on the complexity of the pattern of the pixel. Recently in 2008, Maddalena and Petrosino [7] proposed an interesting biologically inspired self organizing approach using artificial neural network for background modeling. A model of the background was made by learning the motion of variations of the background and based on this learnt background model, the algorithm could extract the region of interest.

The original GMM cited in many literatures assumed that the variances of the red, green and blue were same and independent of each other. Tang and Miao proposed an algorithm to improve the preciseness of the method by considering that each channel has its own variances [8].

In [9] they have proposed an approach for removing false motion detection to a great extent by combining the conventional adaptive GMM with neighborhood based differences and overlapping based classification.

The rest of the paper is organized as follows: In section 3 the Stauffer and Grimson method [1] for GMM shall be explained. In section 4, the method for shadow removal is discussed followed by the proposed approach in section 5. In Section 6 we have the results and discussion and finally the applicability of our work is discussed in section 7.

## III. GAUSSIAN MIXTURE MODEL

A Gaussian Mixture Model (GMM) is a parametric probability density function represented as a weighted sum of Gaussian component densities. GMMs are commonly used as a parametric model of the probability distribution of continuous measurements or features in a biometric system, such as vocal-tract related spectral features in a speaker recognition system. GMM parameters are estimated from training data using the iterative Expectation Maximization (EM) algorithm or Maximum A Posteriori (MAP) estimation from a well-trained prior model.

A Gaussian mixture model is a weighted sum of M component Gaussian densities as given by the equation,

$$p(X|\lambda) = \sum_{i=1}^{M} w_i g(X|\mu_i, \Sigma_i)$$

where x is a D-dimensional continuous-valued data vector (i.e. measurement or features), $w_i$, i=1, . . . ,M, are the mixture weights, and $g(x|\mu_i, )$, i = 1, . . . ,M, are the component Gaussian densities. Each component density is a D-variate Gaussian function of the form,

$$g(X|\mu_i, \Sigma_i) = \frac{1}{(2\pi)^{\frac{D}{2}} |\Sigma_i|^{1/2}} \exp\left\{-\frac{1}{2}(X-\mu_i)'\Sigma_i^{-1}(X-\mu_i)\right\}$$

with mean vector $\mu_i$ and covariance matrix $\Sigma_i$. The mixture weights satisfy the constraint that $\sum_{i=1}^{M} w_i = 1$.

### A. Implementation

A mixture of K Gaussians is used to model the time series of values observed at a particular pixel. The probability of occurrence of the current pixel value is given by,

$$P(Z) = \sum_{i=1}^{K} w_i N(\mu_t, \Sigma_t, Z)$$

Where N is the Gaussian probability density function, whose mean vector is $\mu$ and covariance is $\Sigma$.
And $w_i$ is the weight of the ith Gaussian such that $\Sigma w_i = 1$. The covariance matrix is assumed to be of the form $\Sigma = \sigma^2 I$ for computational reasons.

### B. Parameter updates

The new pixel value $Z_t$ is checked against each Gaussian. A Gaussian is labeled as matched if

$$\|Z - \mu_h\| < d\sigma_h$$

Then its parameters may be updated as follows:

$$w_{i,t} = (1-\alpha) * w_{i,t-1} + \alpha * M_{i,t}$$
$$\mu_t = (1-\rho) * \mu_{t-1} + \rho * Z_t$$
$$\sigma_t^2 = (1-\rho) * \sigma_{t-1}^2 + \rho * (Z_t - \mu_t)^T * (Z_t - \mu_t)$$
$$\rho = \alpha * N(\mu_t)$$

Where $\alpha$ is the learning rate for the weights.
If a Gaussian is labeled as unmatched only its weight is decreased as

$$w_{i,t} = (1-\alpha) * w_{i,t-1}$$

If none of the Gaussians match, the one with the lowest weight is replaced with Zt as mean and a high initial standard deviation.

The rank of a Gaussian is defined as w/σ. This value gets higher if the distribution has low standard deviation and it has matched many times. When the Gaussians are sorted in a list by decreasing value of rank, the first is more likely to be background. The first B Gaussians that satisfy (1) are thought to represent the background.

$$B = \arg\min_b \left( \sum_{k=1}^{b} w_i > T \right) \quad (1)$$

The Gaussian mixture model (GMM) is adaptive; it can incorporate slow illumination changes and the removal and addition of objects into the background. Further it can handle repetitive background changes like swaying branches, a flickering computer monitor etc. The higher the value of T in (1), the higher is the probability of a multi-modal background.

*C. Shadow Removal*

Shadows detected as foreground can cause several problems when extracting and labeling objects, two examples are object shape distortion and several objects merging together. It is especially crucial and these problems should be avoided.

Generally the image is first transformed to a color space that segregates the chromaticity information from the intensity. In the HSV color space, the hypothesis is that a shadowed pixel value's value and saturation will decrease while the hue remains relatively constant. In CIELAB, the luminance component should decrease and the chromaticity coordinates should remain relatively constant.

However, the color model of Horpresert et al [10] gave the best results in the test conducted. It doesn't require any complicated conversion formulae like in the case of HSV and CIELAB. Also the simple choice of parameters is a distinct advantage.

In this model each pixel value in the RGB color space is assumed to lie on a chromaticity line, which connects the pixel value and the origin. The authors specify a way to calculate the deviation of a foreground pixel value from the background value. The foreground value is compared with the means of each of the B background Gaussians in equation (1).

The brightness distortion (BD) and chromaticity distortion (CD) are defined as (see Fig.1):

$$BD = \frac{OP}{OB} \quad \& \quad CD = PF$$

Shadow points (sp) can now be ascertained as:

$$sp = \begin{cases} 1 & for\ \alpha \leq BD \leq \beta \wedge CD \leq \tau_C \\ 0 & otherwise \end{cases}$$

In Fig. 1, OB is the background RGB vector and OF is the foreground RGB vector. FP is the perpendicular dropped from F onto OB. The shaded cylinder is the locus of all shadow color values.

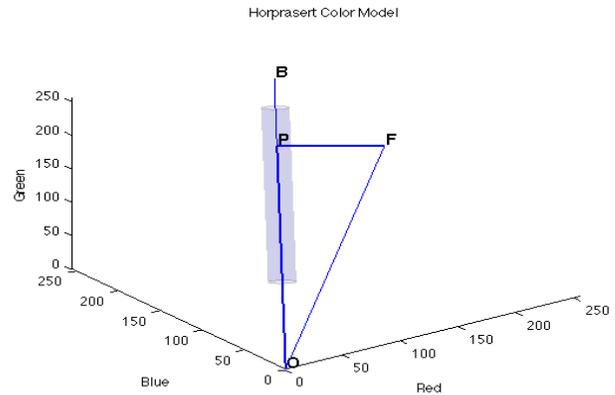

**Figure1:** Brightness and chromaticity distortion in the RGB color space [10].

### IV. PROPOSED APPROACH

We propose an approach to extract the region of interest by using an adaptive GMM discussed above for background subtraction. The shadows are then efficiently detected employing the Horpresert color model as discussed in section 4. After these, when the foreground pixels are identified they can be segmented using the connected component analysis. This would enable us to detect the moving object not only by their positions but we can also get information regarding their size and other shape information and hence this would enhance tracking. We propose to use this for detection of abandoned object in the scene. As our background subtraction algorithm is adaptive, any foreground object that remains static beyond a certain time, becomes a part of background. This can be used for detecting an abandoned luggage. If any motion is detected initially and thereafter the background becomes constant and remains unmoved for next few minutes, it could be possible that an object had been kept in the scene and has not been moved for a long time. This could lead to some suspicious situation and hence an alarm could be triggered.

### V. RESULTS AND DISCUSSION

Fig.2 shows the result of background subtraction using AGMM. Fig.3 shows the results of shadow detection. The figure shows the actual scene accompanied by the result of the segmentation. The foreground pixels are red, whereas the shadow pixels are green. The results obtained using Horpresert model is satisfactory. Also experimentation shows that the discussed model does not require any trained background model, it can adaptively operate under any given set of environment, and is robust to illumination changes and fast repetitive motion. The camera used in the tests is a lenovo G460 easy camera configured with a resolution of 120 X 160 pixels running in a 32 bit operating system, 2.00 GHz processor, and 2 GB RAM. The achieved frames per second of the proposed model is 20 fps.

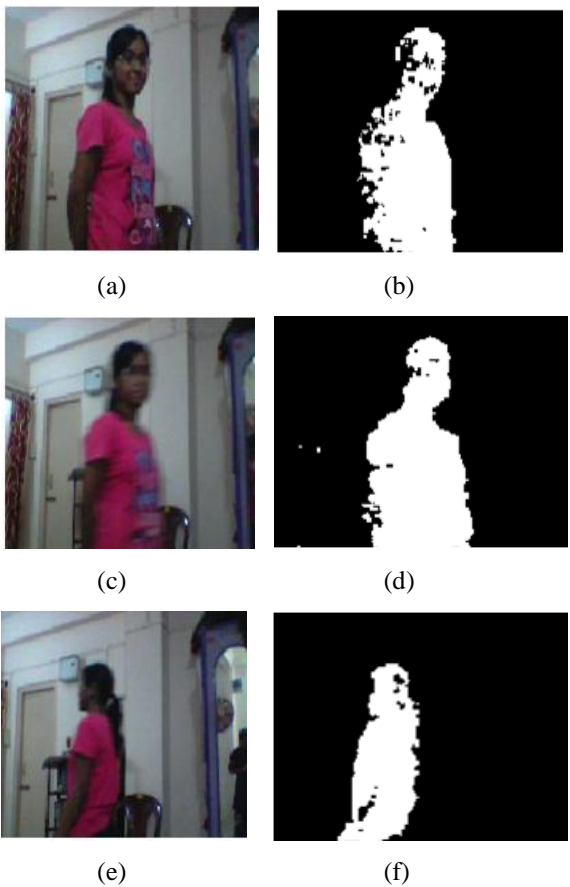

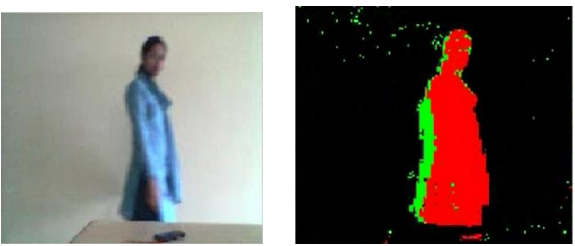

**Figure 2:** (a), (c), (e) are the original frames and (b), (d) & (f) are the segmented foreground using AGMM after background subtraction.

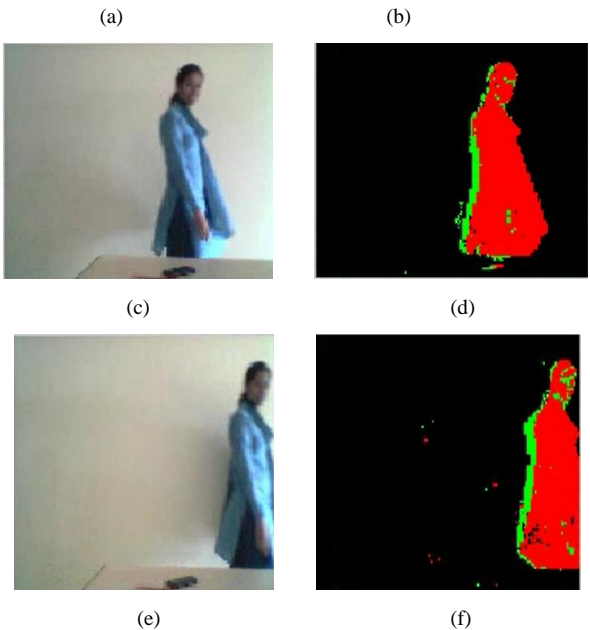

**Figure 3:** (a), (c), (e) are the original frames and (b), (d) & (f) shows the result after background subtraction and shadow detection

## VI. APPLICABILITY

The background subtraction method as discussed can be employed for a good number of applications especially in surveillance. It could be used to detect abandoned luggage in airport and railway platforms and in any place where security is of prime concern. Also it could be used in areas where entry of human being is highly restricted. Instead of employing a human-being to observe such areas, the proposed method could be employed so that any movement in that area can be immediately detected and an alarm can be triggered.